# HySenSe: A Hyper-Sensitive and High-Fidelity Vision-Based Tactile Sensor

Ozdemir Can Kara[1], Naruhiko Ikoma[2], and Farshid Alambeigi[1]
[1]Walker Department of Mechanical Engineering, University of Texas at Austin, Austin, TX, USA
[2]Department of Surgical Oncology, MD Anderson Cancer Center, Houston, TX, USA
farshid.alambeigi@austin.utexas.edu

*Abstract*—In this paper, to address the sensitivity and durability trade-off of Vision-based Tactile Sensor (VTSs), we introduce a hyper-sensitive and high-fidelity VTS called HySenSe. We demonstrate that by solely changing one step during the fabrication of the gel layer of the GelSight sensor (as the most well-known VTS), we can substantially improve its sensitivity and durability. Our experimental results clearly demonstrate the outperformance of the HySenSe compared with a similar GelSight sensor in detecting textural details of various objects under identical experimental conditions and low interaction forces (≤1.5 N).

## I. INTRODUCTION

Amid the tactile sensors, Vision-based Tactile Sensors (VTSs) have recently been developed to improve tactile perception via high-resolution visual information [1]. VTSs can provide high-resolution 3D visual image reconstruction and localization of the interacting objects by capturing tiny deformations of an elastic gel layer that directly interacts with the objects' surface [2]. GelSight is the most well-known VTS, developed by Johnson and Adelson [3], and has been utilized for various applications, including the surface texture recognition [4], geometry measurement with deep learning algorithms [5], localization and manipulation of small objects [6], and hardness estimation [7]. More details about the use of VTSs can be found in [8].

The resolution and quality of the GelSight output (i.e., 3D images) highly depends on its hardware components (e.g., the utilized elastomer, optics, illumination [5], [9]), fabrication procedure (e.g., thickness and hardness of gel layer [10]), and post-processing algorithms [11]. More importantly, the above-mentioned parameters also directly affect the *sensitivity* and the *durability* of the fabricated VTS. It is worth mentioning that, in this study, *sensitivity* is defined as the ability to obtain high-quality 3D images while applying low interaction forces independent of the shape, size, and material properties of objects, whereas *durability* refers to the effective life of the VTS without experiencing any wear and tear after multiple use cases on different objects.

The sensitivity and durability of VTSs are highly correlated. To build a sensitive sensor using the fabrication procedure proposed in [10], durability needs to be often compromised

*Research reported in this publication was supported by the University of Texas at Austin and MD Anderson Cancer Center Pilot Seed Grant.

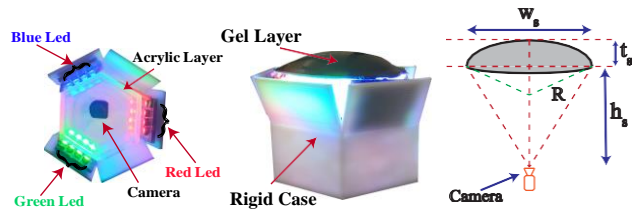

Fig. 1: Constructing elements of a VTS.

and vice versa. For instance, reducing the gel layer's stiffness and/or thickness is a common technique to increase the sensitivity of the GelSight sensors; however, this approach may substantially reduce the durability of the sensor when interacting with different objects [5], [11]. Moreover, to obtain a high-resolution image using a less sensitive GelSight sensor (i.e., having a high thickness and stiffness gel layer), often a higher interaction force is required to deform the gel layer and obtain high-resolution images. Of note, this might not be feasible for several applications (e.g., high-fidelity manipulation of fragile objects [12] and surgical applications [13]–[15] and may damage the sensor and reduce its durability. Therefore, there is a critical need for developing a VTS that simultaneously has high sensitivity and durability independent of its application.

To address the sensitivity and durability trade-off of common VTSs, in this paper, we present the design and fabrication of a hyper-sensitive and high-fidelity VTS (called *HySenSe*) that requires a very low interaction force (i.e., < 1.5$N$) to obtain high-resolution images. To fabricate HySenSe, we follow the standard fabrication procedure of GelSight sensors [10] and show that with a minor change in the fabrication procedure, (i) we can drastically improve its sensitivity and obtain high-quality images, (ii) while applying a very low interaction force that does not compromise its durability. To thoroughly evaluate the performance of HySenSe, we analyze and compare its 3D image outputs with a similar Gelsight sensor on different objects (with different shapes, textures, and stiffness) and under various interaction forces.

## II. MATERIALS AND METHOD

*1) Working Principle and Constructing Elements of HySenSe:* As illustrated in Fig. 1 and similar to the GelSight sensor [3], HySenSe consists of a dome-shape deformable silicone layer that directly interacts with an object, a camera that faces toward the gel layer and captures the deformation of the gel layer, and is fixed to the rigid frame of the sensor, a transparent acrylic layer that supports the gel layer and an array of Red, Green and Blue LEDs creating illumination and aiding in a recreation of the 3D textural features when an object interacts with the sensor. The *working principle* of GelSight and HySenSe are identical and very simple yet highly intuitive. The deformation caused by the interaction of the gel layer with the object can be visually captured by the small camera embedded in the frame.

*2) Fabrication Procedure:* To thoroughly evaluate the performance of HySenSe, we first fabricated similar GelSight and HySenSe sensors. For both sensors, we used a 5 MP camera (Arducam 1/4 inch 5 MP sensor mini) that was fixed to the rigid frame printed with high-resolution printing technology, a Form 3 printer (Formlabs Form 3, Formlabs Inc.), and the clear resin material (FLGPCL04, Formlabs Inc.). The rigid frame height was designed as $h_s$=24 mm, determined based on the camera focus and field of view. Of note, the other dimensions of the rigid frame are determined based on the size of the gel layer described below. Moreover, an array of Red, Green and Blue LEDs (WL-SMTD Mono-Color 150141RS63130, 150141GS63130, 150141BS63130, respectively) were placed and arranged 120 degrees apart. To have identical fabricated gel layers for these sensors and in order to build samples with identical volume and geometry, we calculated the volume of the spherical shape gel layers V as follows:

$$V = \frac{1}{3}\pi t_s^2(3R - t_s) \quad (1)$$

where, as conceptually demonstrated in Fig. 1, $w_s$ is the width of the gel layer, $t_s$ is the thickness of the fabricated samples, $h_s$ is the height of the rigid frame, and $R$ is the radius of the hemispherical-shape gel layer.

*Fabrication Procedure of the Gel Layer:* As mentioned, the sensitivity of a VTS can be controlled using the utilized hardware components (e.g., [10]) and/or post processing algorithms (e.g., [11]). Nevertheless, in this study, to fabricate a hyper-sensitive and durable VTS, we are showing that solely by changing one step during fabrication of the GelSight's gel layer (as suggested in [5]), we can substantially increase its sensitivity and durability. The following briefly describes and compares the steps proposed for fabricating a gel layer for GelSight sensor and our proposed modified procedure for fabricating the gel layer for the HySenSe sensor.

**GelSight's Gel Layer:** <u>STEP 1:</u> For the fabrication of the gel layer (shown in Fig. 2), we first used a soft transparent platinum cure two-part (consisting of Part A and Part B) silicone (P-565, Silicones Inc.). In this study, we used 14:10:4 (A:B:C) as a mixture mass proportion, in which Part C corresponds to phenyl trimethicone- softener (LC1550, Lotioncrafter). Notably, this proportion can readily be changed depending on the application requirements. Next, to fabricate a hemispherical-shape gel layer (shown in Fig. 1), we used a silicone mold (Baker depot mold for chocolate) with $R$= 35 mm and coated its surface with Ease 200 (Mann Technologies) to prevent adhesion and ensure a high surface quality after molding. After the coating was dried, the prepared silicone mixture was poured into the mold and then degassed in a vacuum chamber to remove the bubbles trapped within the mixture. Next, samples were cured in a curing station (Formlabs Form Cure Curing Chamber).

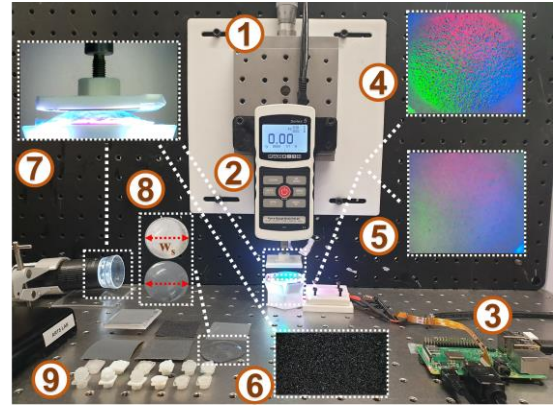

Fig. 2: Experimental Setup: 1- Linear Stage, 2- Force Gauge, 3- Raspberry Pi, 4- HySenSe's image output of the sandpaper, 5- GelSight's image output of the sandpaper (2.5 N), 6- 150 grit sandpaper, 7- Side View of the interaction surface, 8- HySenSe (above) and GelSight (below) gel layers, $w_s$ = 33.8 mm, 9- Different objects used for the experiments.

<u>STEP II</u>: After curing, the matte-colored aluminum powder (AL-101, Atlantic Equipment Engineers) was brushed on the gel layer's dome surface to avoid light leakage. Of note, the fabricated gel layer had $w_s$= 33.8 mm and $t_s$= 4.5 mm, which relatively has a higher thickness compared to the previous literature (e.g., [5], [11].). Remarkably, these parameters can readily be changed depending on the application requirements.

<u>STEP III</u>: Finally, a thin layer of silicone with the addition of grey pigment (blend of both black and white pigments- Silc Pig Black, Silc Pig White, Smooth-On Inc) was poured, with identical proportion described in STEP I, on the surface of gel layer to prevent light leakage and stabilize the aluminum powder layer. Notably, the hardness of the fabricated gel layer was measured as 00-20 using a Shore 00 scale durometer (Model 1600 Dial Shore 00, Rex Gauge Company). The fabricated gel layer is shown in Fig. 2.

**HySenSe's Gel Layer:** To fabricate the gel layer for HySenSe sensor (shown in Fig. 2), we exactly followed the above-described procedure in *STEP I*. The major change in the proposed HySenSe fabrication procedure happens in *STEP II* in which, instead of brushing the aluminum powder on the surface of the fabricated gel layer, we utilized Specialty Mirror Effect spray paint (Rust-Oleum Inc.) and sprayed the surface of fabricated gel layer for 5 times with 1-minute intervals (to ensure the sprayed paint is cured). Using the spray paint eliminates the need for the addition of grey pigment with the silicone mixture in *Step III*. Thus, as the last fabrication step, a thin layer of silicone mixture, prepared in *Step I*, was poured on the surface of the gel layer to cover the spray coating. Notably, the hardness of the fabricated gel layer was measured as 00-18 using the Shore 00 scale durometer.

## III. EXPERIMENTAL PROCEDURE AND RESULTS

Fig. 2 shows the experimental setup used to evaluate the sensitivity and fidelity of the HySenSe and GelSight sensors by comparing their textural images while measuring the interaction forces between their gel layers and the objects. As shown, the setup consists of HySenSe and GelSight sensors, a single-row linear stage with 1 $\mu m$ precision (M-UMR12.40, Newport), a digital force gauge with 0.02 N resolution (Mark-10 Series 5, Mark-10 Corporation) attached to the linear stage to precisely push various objects on the sensors gel layers and measure the applied interaction force, and a Raspberry Pi 4 Model B for streaming and recording the obtained images by sensors. We also utilized MESUR Lite Basic data acquisition software (Mark-10 Corporation) to record the interaction forces between the gel layers and objects.

To thoroughly compare the sensitivity and durability of the HySenSe and GelSight sensors independent of the size, shape, thickness, texture, and hardness of objects, as shown in Fig. 3, we considered distinct test cases. To evaluate performance of the sensors on flat and thin objects with different hardness and texture, we made samples from a 150 grit silicon carbide sandpaper (SKOCHE) with 95 $\mu m$ grain size and a piece of soft paper towel (Bounty, Procter & Gamble Inc.) with 1.1 $mm$ textural details. Also, to investigate the sensitivity of fabricated sensors on objects with non-flat geometry, heterogeneous texture, and distinct material hardness, we 3d printed soft (type LST, DM400 material) and hard (type IIc, Vero PureWhite material) colorectal polyp phantoms (with Shore hardness 00-45, D-83, respectively) based on the Paris classification [16], using Digital Anatomy Printer (J750, Stratasys, Ltd) [17]). Dimensions (length × width × height) of the soft and hard polyp phantoms are 10.21 × 9.69 × 3.16 $mm$ and 12.05 × 12.00 × 1.11 $mm$, respectively. For the performed experiments, we first attached and secured these samples to the force gauge. Then, the gel layers were fixated on the rigid frame, and the whole sensor was fixed to the optical table to block any undesired motion. Next, the linear stage was precisely pushed on the sensors' gel layers while measuring the displacement of the linear stage (i.e., deformation of the gel layer), interaction force, and obtained images by the sensors. Fig. 3 summarizes the results of the performed experiments.

## IV. DISCUSSION AND CONCLUSION

To improve the resolution of the VTS image outputs, various approaches such as increasing the interaction forces (e.g., [7]), reducing the hardness and thickness of gel layer, complex fabrication procedures (e.g., [5]), and post processing image processing algorithms (e.g., [11]) have been implemented and proposed in the literature. Nevertheless, we demonstrated that by solely changing one step (i.e., *STEP II*) during fabrication procedure of the gel layer, the sensitivity of the GelSight sensor can be drastically improved. More specifically, by using a mirror spray paint instead of aluminium powder and grey pigments, not only we can improve the reflectivity of the illumination, but we also can reduce the thickness of the coating to substantially improve the sensitivity.

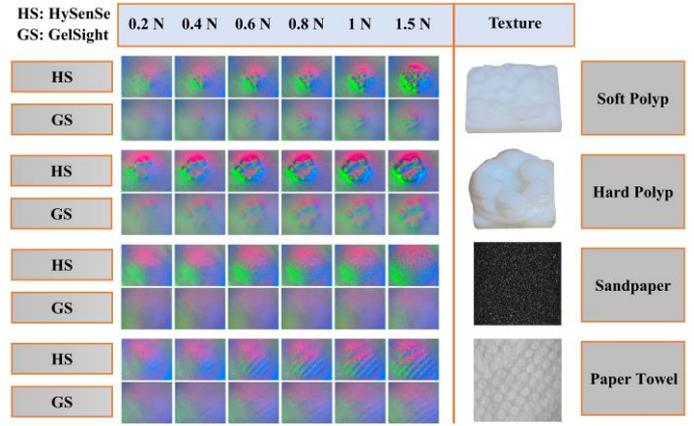

Fig. 3: Evolution of the visual outputs for the HySenSe and GelSight sensors. Each two rows of the figure corresponds to a specific object used for experiments. Also, the top row indicates the applied forces corresponding to each image.

Fig. 3 clearly demonstrates the superior sensitivity of the HySenSe compared with the GelSight sensor obtained under identical experimental conditions. As can be observed in this figure, the HySenSe sensor demonstrates a substantially better performance than the GelSight sensor in creating high-fidelity images for all of the used objects independent of their hardness, size, thickness, and texture at very low interaction forces (i.e., ≤ 1.5 N). Particularly, even at ≤ 0.6 N interaction force, HySenSe can provide very visible and high-quality textural images (e.g., 95$\mu m$ grains in the sandpaper), whereas at these low forces, GelSight's outputs are very blurry and unclear. Of note, this important feature is critical for several applications (e.g., high-fidelity manipulation of fragile objects [12] and medical applications [13]) in which a high sensitivity at low interaction forces is necessary to ensure the safety of tactile measurements while providing a high-resolution image output.

Aside from sensitivity, the proposed fabrication procedure for the HySenSe also can resolve the existing sensitivity and durability trade-off in VTSs. As described, to improve the durability of the GelSight sensors, the thickness and/or the hardness of the gel layer needs to be increased while this may drastically deteriorate the sensitivity of these sensors [9]. The typical remedy for such situations is to increase the interaction force between the object and the gel layer to obtain a high-quality image, which may not be feasible for many applications and may damage the sensor, too. Nevertheless, as shown in Fig. 3, the hypersensitivity of the HySenSe compared with the GelSight sensor, even in very low interaction forces, addresses the issue of sacrificing the durability for sensitivity and vice versa. In other words, the hypersensitivity of HySenSe mitigates the need for applying a high interaction force that may reduce the durability and the effective life of the sensor. Instead of a qualitative comparison of the HySenSe and GelSight image outputs, in the future, we plan to use quantitative image comparison metrics (e.g., [18]) to represent the out-performance of the proposed sensor on various objects with different textures, sizes, hardness, and materials.